%% file: root.tex
\documentclass[letterpaper, 10 pt, conference]{ieeeconf}  % Comment this line out if you need a4paper

\IEEEoverridecommandlockouts                              % This command is only needed if 
\usepackage{soul}
\usepackage{dingbat}
\usepackage{tabularx}
\usepackage{multirow}
\usepackage{siunitx}
\usepackage{booktabs}
\usepackage{adjustbox}
\usepackage{amsmath}
\usepackage{bm}
\usepackage{amsfonts} 
\usepackage{cite}
\usepackage{pifont}% http://ctan.org/pkg/pifont

\newcommand{\NA}{---}
\newcommand{\pluseq}{\mathrel{+}=}

\newcommand{\leftcelll}[1]{\multicolumn{1}{l|}{#1}}
\newcommand{\leftcell}[1]{\multicolumn{1}{l}{#1}}

\usepackage[dvipsnames]{xcolor}

\title{\LARGE \bf
Visual Forecasting as a Mid-level Representation for Avoidance}

\author{Hsuan-Kung Yang$^{1}$, Tsung-Chih Chiang$^{1}$*, Ting-Ru Liu$^{1}$*, Chun-Wei Huang$^{1}$*, Jou-Min Liu$^{1}$*, and Chun-Yi Lee$^{1}$% <-this % stops a space
\thanks{* indicates equal contribution.}% <-this % stops a space
\thanks{$^{1}$ Elsa Lab, Department of Computer Science, National Tsing Hua University, Hsinchu City, Taiwan.}
\thanks{This work has been submitted to the IEEE for possible publication. Copyright may be transferred without notice, after which this version may no longer be accessible.}
}

\begin{document}

\maketitle
\thispagestyle{empty}
\pagestyle{empty}

%%%%%%%%%%%%%%%%%%%%%%%%%%%%%%%%%%%%%%%%%%%%%%%%%%%%%%%%%%%%%%%%%%%%%%%%%%%%%%%%
\begin{abstract}
The challenge of navigation in environments with dynamic objects continues to be a central issue in the study of autonomous agents. While predictive methods hold promise, their reliance on precise state information makes them less practical for real-world implementation. This study presents visual forecasting as an innovative alternative. By introducing intuitive visual cues, this approach projects the future trajectories of dynamic objects to improve agent perception and enable anticipatory actions. Our research explores two distinct strategies for conveying predictive information through visual forecasting: (1) sequences of bounding boxes, and (2) augmented paths. To validate the proposed visual forecasting strategies, we initiate evaluations in simulated environments using the Unity engine and then extend these evaluations to real-world scenarios to assess both practicality and effectiveness. The results confirm the viability of visual forecasting as a promising solution for navigation and obstacle avoidance in dynamic environments.
\end{abstract}

%%%%%%%%%%%%%%%%%%%%%%%%%%%%%%%%%%%%%%%%%%%%%%%%%%%%%%%%%%%%%%%%%%%%%%%%%%%%%%%%
\section{INTRODUCTION}
The challenge of navigation and obstacle avoidance in environments containing dynamic objects has been a central focus in the field of autonomous agents. One prevailing direction to address this challenge involves utilizing predictive information to augment the performance of the agents~\cite{LIU2019272, poddar2023crowd, wang22e, Park_2016_IROS, Chen2018RobotNB, Fraichard2020FromCS, trajectory-pred-for-avoidance, Mavrogiannis2021WindingTC}. While the predictive methods have shown to improve performance when applied to downstream tasks, a notable limitation lies in their dependence on accurate state information for both the agent and the environment. Obtaining the requisite data often demands considerable effort and is therefore a primary hurdle in implementing these methods for real-world obstacle avoidance and navigation tasks. In addition, the complex interaction and dynamics in such scenes make it difficult for agents to interpret, respond to, or model outcomes effectively and efficiently. As a result, these difficulties have prompted interests in exploring alternative approaches that rely on more direct, visually-oriented representational solutions that can offer an immediate and intuitive understanding of environmental dynamics.  Within this context, we introduce the concept of visual forecasting as a compelling solution.  This methodology utilizes intuitive visual cues to represent the future trajectories of dynamic objects and serves dual purposes: facilitating the perception process for the agent and augmenting its ability to take anticipatory action and respond effectively to prospective events. Moreover, this methodology obviates the necessity for precise acquisition of environmental states, which offers promising avenues for implementation in real-world settings.

In light of the above, it becomes essential to identify an approach to incorporate visual forecasting information into the representations presented to an autonomous agent in a manner that is both effective and efficient.  The task entails the translation of complex dynamic information into a format of visual observation that is digestible and comprehensive, enabling the agent to interpret and react promptly.   Overly simplified or exceedingly complicated representations could result in inefficient processing or the omission of vital cues. A viable solution to this challenge may be grounded on the adoption of mid-level representations~\cite{Hong2018IJCAI, zhao2020sim2real, Lin2020LearningTS, chen_2020_CORL}. Such a concept offers the potential to achieve a balance between high-level abstract information and low-level sensory data. By utilizing mid-level representations, the visual forecasting strategies concerned in this study can encapsulate essential aspects of forecast information in a manner that is sufficiently comprehensive and computationally efficient. This study embarks on an investigation of two potential strategies for representing visual forecasting through mid-level representations: (1) a sequence of bounding boxes, and (2) an augmented path. Each strategy presents a unique mechanism for modifying and integrating information into mid-level representations so as to convey future predictive information to the agent. Our objective is to investigate whether these modified mid-level representations can enhance comprehension and provide feasible guidance to the autonomous agents.

\begin{figure}[t]
  \centering
  \includegraphics[width=\linewidth]{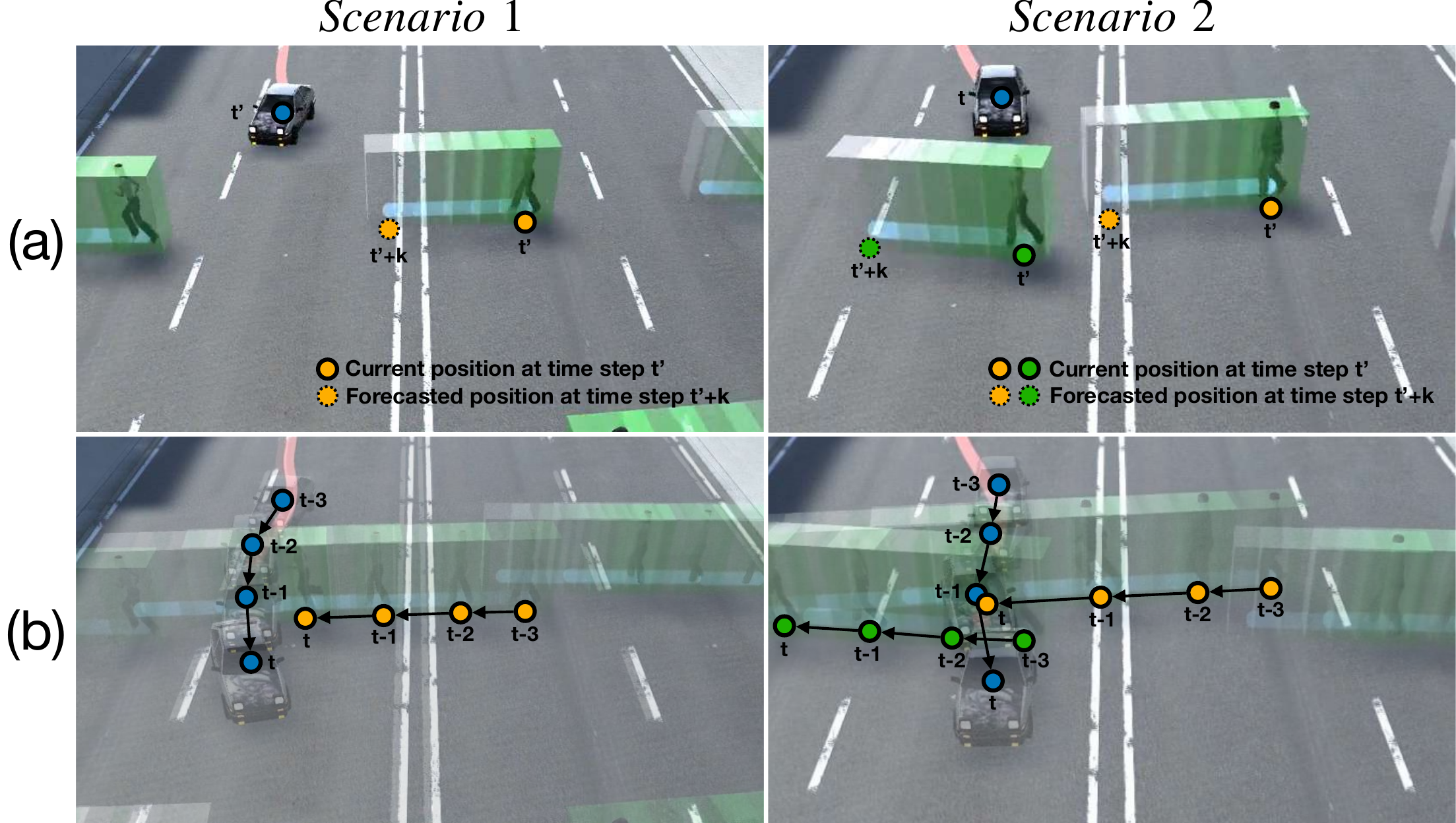}
  \caption{The demonstrations of (a) the forecasted trajectories and (b) the agents interacting and avoiding pedestrians in the simulated environments.}
  \label{fig:exp-virtual}
\end{figure}

To validate the proposed visual forecasting strategies, we first utilize the Unity engine~\cite{unity-eng} to conduct a comprehensive set of evaluations in simulated environments. These environments enable precise generation, configuration, and tracking of dynamic object movements, and offer complete customization capabilities for rigorous evaluations. Our assessment begins by examining the effectiveness of visual forecasting and then proceeds to analyze the underlying causes of the failure cases. Moreover, our evaluations explore the impact of forecasting quality on the performance of downstream tasks, specifically in object avoidance tasks performed by deep reinforcement learning (DRL) agents. To augment the validation and extend the applicability of our concepts, real-world scenarios are incorporated into this study. While these practical scenarios inherently contain challenges such as complexity and uncertainty, they provide valuable insights into the feasibility, adaptability, and practicality of our proposed forecasting schemes. From both simulated and real-world settings, our analyses and findings serve to validate the effectiveness of visual forecasting in assisting autonomous agents. Furthermore, they highlight both its practical viability and potential for broader applications.

\section{Preliminary}
\label{sec:preliminaries}
\subsection{Virtual-to-Real Transfer via Mid-Level Representations}
\label{subsec::virtual_to_real_mid_level}
Mid-level representations serve as crucial abstract constructs that encapsulate a variety of physical or semantic aspects inherent in visual scenes. These domain-invariant properties, whether extracted or inferred, have demonstrated their significance across a broad spectrum of applications, and facilitate efficient information transfer from perception modules to control modules~\cite{Hong2018IJCAI, zhao2020sim2real, Lin2020LearningTS, chen_2020_CORL, yang2023vision}. These representations can take a variety of forms, such as depth maps, raw optical flow, semantic segmentation, etc. Each of these forms exhibits unique strengths and potential limitations depending on the specifics of the scenario~\cite{Yang2022IROS}. Recent research~\cite{yang2023vision}  has introduced the concept of virtual guidance as a novel mid-level representation. This approach generates semantic segmentation-like virtual markers to direct agents along a predetermined path. This reveals the potential of adaptively modifying mid-level representations as a means to supply agents additional information to enhance their decision-making abilities. Recognizing the significance of mid-level representations is essential for the successful implementation of modular learning-based frameworks.  As a result, this study aims to utilize the concept of mid-level representation and explore methods to represent forecasted trajectories of dynamic objects to facilitate decision-making of the agents.

\subsection{From Forecasting to Action}
The primary focus of several seminal works has predominantly been on the aspect of prediction, with less emphasis placed on utilizing these predictive models for subsequent downstream tasks~\cite{Gupta_2018_CVPR, Salzmann_2020_TrajectronDT, Mangalam_2020_WACV, Yuan_2021_ICCV, Pang_2021_CVPR, Shafiee_2021_CVPR}.  Nonetheless, recent literature increasingly emphasizes the advantageous integration of forecasting models into various tasks. For instance, the study in~\cite{unsup-interaction-video-prediction} explored the domain of visual forecasting with action-conditioned predictions, which can be used for futuer action planning. Within the RL domain, PiSAC~\cite{PiSAC} highlighted the advantages of predictive information, and suggested that the incorporation of such information can expedite the learning process.  In the context of navigation tasks, several works have made progress.  Studies such as~\cite{Fraichard2020FromCS, trajectory-pred-for-avoidance, Mavrogiannis2021WindingTC} have combined forecasting with model predictive control (MPC) to enhance performance. However, these studies often rely on the assumption that position data are readily available, and oftentimes utilize straightforward location information instead of richer high-dimensional RGB data. Deriving such positional information, however, typically necessitates additional sensors or sophisticated position estimation models. There have been research endeavors~\cite{visualforesight, Zeng_2020_CVPR} which have integrated RGB input with MPC and demonstrated the viability of RGB-based control tasks. Despite the feasibility of MPC for forecasting in a range of tasks, applying this technique to real-world visual navigation and obstacle avoidance remains challenging. One issue that warrants mention is the need for an accurate world model. Such models often face difficulties when dealing with high-dimensional state spaces inherent in visual inputs. It is important to clarify that this study aims to introduce methods for representing forecasted information on visual observations. Specifically, the focus is on presenting information related to dynamic entities in a manner that is both intuitive and easily comprehensible for agents. While it holds potential for integration with MPC, the objective of this paper remains distinct from MPC based methodologies.

%===============================================================================

\section{Methodology}
\label{sec:methodology}
Visual forecasting is a representational approach that aims to abstract and illustrate forecasted information about moving objects for an agent to process such information as visual inputs. Visual forecasting can serve as a type of mid-level representation and is well-suited for both virtual and real settings as well as virtual-to-real transfer learning. This is largely due to its capacity to express forecasted data, such as future trajectories of dynamic objects, in clear and readily interpretable formats using basic shapes and colors. At its core, visual forecasting serves the purpose of translating forecasted information into visual perceptions. This approach carries multiple advantages. The primary benefit lies in facilitating intuitive comprehension.  By rendering temporally forecasted information into visually understandable formats, visual forecasting simplifies the processing requirements for the agent.  Second, it enhances the agent's grasp of contextual information. While observations made by the agent typically contain information from only the current timestep, the modification of mid-level representation to incorporate visual forecasting enables the agent to access future forecasted information within its current observation. By enabling the agent to directly observe the forecasted trajectories of dynamic objects, it can more holistically perceive the environment and its dynamics, thereby supporting more informed decision-making.  We next introduce a framework along with two visual forecasting schemes to validate both the effectiveness and the feasibility of the proposed concept.

\begin{figure*}[t]
  \centering
  \includegraphics[width=\textwidth]{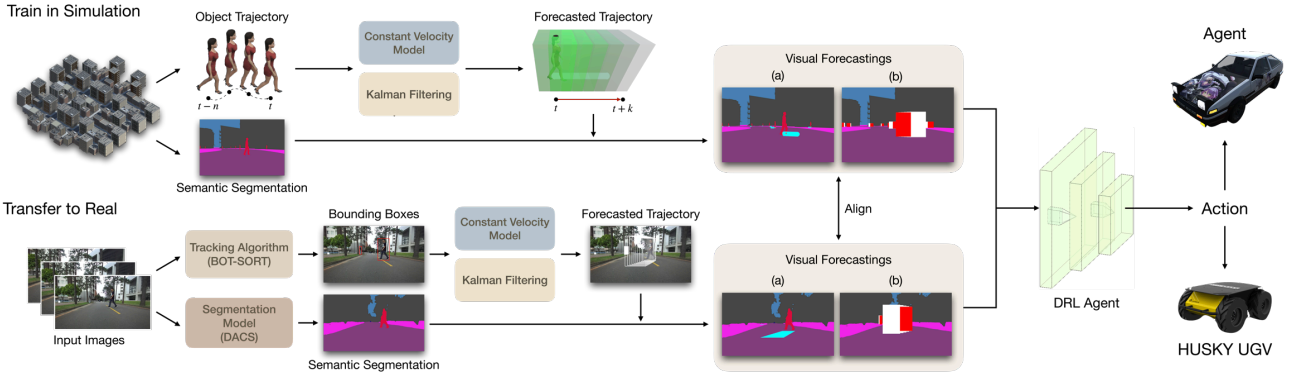}
  \caption{An Overview of our framework.}
  \label{fig:overview}
\end{figure*}

\subsection{Overview of the Framework}
\label{subsec:overview}
Fig.~\ref{fig:overview} provides an overview of the framework, which is composed of two parts: (1) \textit{Train in Simulation} and (2) \textit{Transfer to Real}.  In the \textit{Train in Simulation} phase, each round of simulation produces a set of mid-level representations, including semantic segmentation, detected bounding boxes, and tracked trajectories. These representations form the basis for estimating future trajectories, which can be derived from either a Constant Velocity Model (CVM) or the Kalman Filter (KF)~\cite{kalmanfilter}. For the \textit{Transfer to Real} phase, input images undergo preprocessing via an object detection and tracking algorithm (e.g., BOT-SORT~\cite{aharon2022bot}) as well as a segmentation model (e.g., DACS~\cite{Tranheden_2021_WACV}). This preprocessing generates mid-level representations that correspond to those used in simulation for predicting future trajectories. This work examines two visual forecasting schemes, detailed in Section~\ref{subsec:type_of_visual_forecasting}. Either of these schemes is rendered directly onto the semantic segmentation to form the agent's input observation. When transferring the trained DRL agents to real-world scenarios, the visual forecasting scheme employed aligns with the one utilized in the simulated environments. The configurations pertaining to visual forecasting, such as the colors and geometries superimposed onto the input observations, are preserved to ensure consistency. The trained DRL agent's model is fixed, eliminating the need for further fine-tuning when transferring from virtual to real-world environments.

\subsection{Representaions for Visual Forecasting }
\label{subsec:type_of_visual_forecasting}
\begin{figure}[t]
  \centering
  \includegraphics[width=\linewidth]{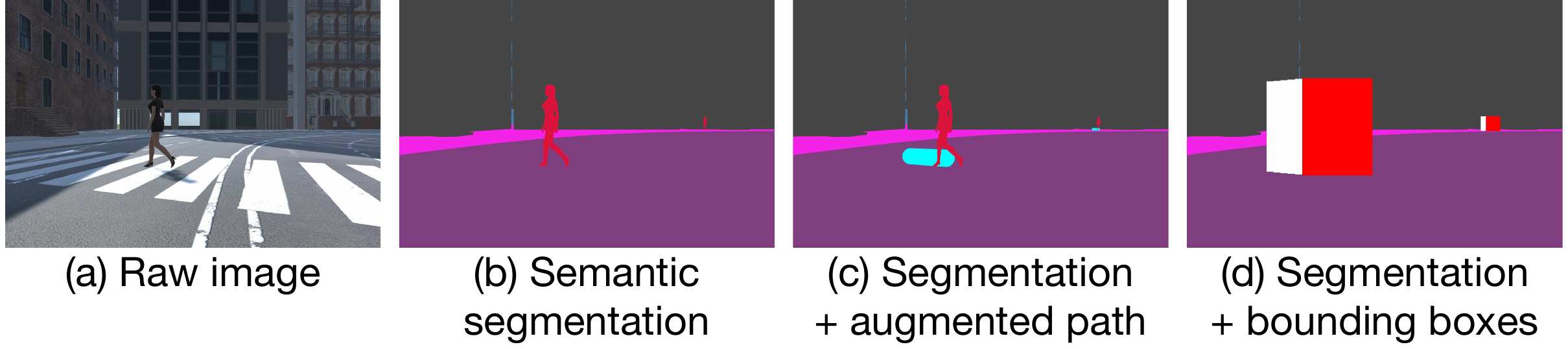}
  \caption{Visualization of different visual forecasting representation schemes.}
  \label{fig:forecasting-visualization}
\end{figure}

In this section, we discuss two visual forecasting strategies utilizing different perspectives on object motion prediction.

\paragraph{Bounding Box (BOX)} 
One potential strategy for visual forecasting can be represented through a sequence of bounding boxes, denoted by ${B_t, B_{t+1}, \ldots, B_{t+k}}$, which indicate the predicted future locations of a specific target object at timesteps ranging from $t$ to $t+k$. Specifically, each bounding box $B_t$ is characterized by the tuple $(x_t, y_t, w_t, h_t)$, where $(x_t, y_t)$ denotes the coordinates of the upper-left corner of $B_t$, whereas $w_t$ and $h_t$ represent the width and height of $B_t$, respectively. This strategy is depicted in Fig.~\ref{fig:forecasting-visualization}~(d).

\paragraph{Augmented Path (AP)}
The augmented path represents another strategy constructed by connecting coordinates from the bounding boxes $B_t$ and $B_{t+k}$. Specifically, the coordinates $(x_{t}, y_{t} + h_{t})$, $(x_{t} + w_{t}, y_{t} + h_{t})$, $(x_{t+k}, y_{t+k} + h_{t+k})$, and $(x_{t+k} + w_{t+k}, y_{t+k} + h_{t+k})$ are linked to create an area that visualizes an object's forecasted trajectory. This strategy offers an intuitive depiction of the future path by extending the lower contours of the bounding boxes, and creates a visual pathway to indicate the anticipated object motion from $t$ to $t+k$. This strategy is depicted in Fig.~\ref{fig:forecasting-visualization}~(c).

\subsection{Generating Visual Forecasting}
\label{subsec:generate_visual_forecasting}
This section presents the generation workflow,  which encompasses two distinct aspects: (1) the derivation of visual forecasting within the simulated environments, and (2) the formulation of visual forecasting in real-world scenarios.

\subsubsection{Derivation of Visual Forecasting in Simulated Environments}
\label{subsubsec:compute_visual_forecasting_in_sim}

Within our simulated environments, historical trajectories of moving objects, specifically pedestrians in this study, are directly obtainable through the Unity engine~\cite{unity-eng}. These trajectories include the complete history of positions and bounding boxes for the moving objects in world coordinates. Given a sequence of pedestrian positions and bounding boxes, either CVM or KF can be employed to estimate future positions and bounding boxes. These estimations then undergo post-processing and are represented via the visual forecasting schemes discussed in Section~\ref{subsec:type_of_visual_forecasting}.  Once these visual forecastings of moving objects are generated, they are rendered to become observable from the agents' viewpoints.

\subsubsection{Formulation of Visual Forecasting in Real-world Scenarios}
\label{subsubsec::forecast_real_world}
Distinct from the procedure discussed in Section~\ref{subsubsec:compute_visual_forecasting_in_sim}, generating visual forecasting in real-world scenarios presents a unique set of challenges. While techniques such as CVM and KF can still be employed to estimate future pedestrian locations, acquiring a complete historical trajectory for each pedestrian could be a complicated matter, especially for a moving camera. The difficulty arises primarily from the ego-centric motion induced by the moving camera. Such motion introduces ambiguity between the object's own motion and the observer's motion, which complicates the task of identifying the cause of bounding box movement. This ambiguity may lead to inaccuracies of state vectors used in KF, and impact the performance of tracking algorithms that rely on KF~\cite{Wojke2017simple, zhang2022bytetrack}. To mitigate the effects of ego-centric motion, BOT-SORT, which incorporates camera-motion compensation, is utilized to deliver more stable historical trajectories. These trajectories, which comprise positions and bounding box coordinates, are utilized as inputs for either CVM or KF to derive forecasted bounding boxes.

%===============================================================================

\label{sec:result}
\begin{figure}[t]
  \centering
  \includegraphics[width=\linewidth]{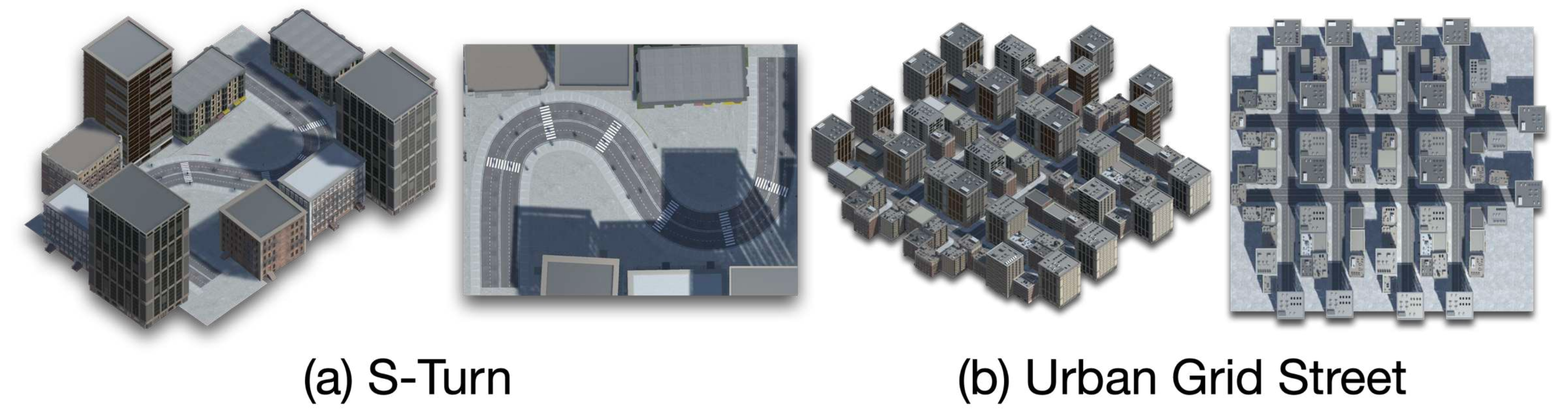}
  \caption{An overview of the simulated environments used in our experiments. The \textit{S-Turn} environment features an S-shaped path, while \textit{Urban Grid Street} is a setting with eight intersections.}
  \label{fig:env}
\end{figure}

\section{Experimental Results}
\label{sec::experimental_results}
The validations of our methodology are conducted in both virtual and real-world settings. In real-world environments, we rely exclusively on a monocular camera for capturing visual data, and no depth sensor (e.g., LiDAR) is utilized.

\subsection{Experimental Setup}
\label{subsec::experimental_setup}

\subsubsection{Virtual Environment Setup}
\label{subsubsec::env_setup}
To evaluate the effectiveness of visual forecasting, we utilize two environments developed using the Unity engine~\cite{unity-eng} in our experiments: \textit{S-Turn}~\cite{Yang2022IROS} and \textit{Urban Grid Street}~\cite{yang2023vision}. Visual depictions of these environments are provided in  Fig.~\ref{fig:env}. The simulated environments employed in our experiments are designed with configurable starting points and destinations, and incorporate dynamic objects (e.g., pedestrians) with customizable speeds. The objective of the agent is to navigate while avoiding both dynamic and static objects, based on the provided observations such as semantic segmentation and visual forecasting.

In the \textit{S-Turn} environment, agents undergo a straightforward test, being trained and evaluated on identical route combinations, albeit with varied pedestrian speeds. The \textit{Urban Grid Street} environment, on the other hand, offers a more rigorous testing scheme. Following the setups in~\cite{yang2023vision}, the routes are categorized into: (a) \textit{seen} and (b) \textit{unseen} routes. The \textit{seen} routes scenario  aims to evaluate the agents' capability to navigate to their destination without encountering obstacles, using the same combinations of starting points and destinations as those in the training phase. In contrast,  the \textit{unseen} routes challenge agents with novel combinations of start and end points.  In total, 89 routes are used for training the agents, while a distinct set of four routes are selected for the evaluation phase. In both environments, pedestrians are configured with speeds of $0.6$ to $\SI{1.2}{\meter/\second}$ during training, and a wider range of $0.3$ to $\SI{1.5}{\meter/\second}$ for the evaluation phase.

\subsubsection{Agent Setup}
In our experiments, the DRL agent is implemented as a deep neural network (DNN) trained using the Soft Actor-Critic (SAC) algorithm~\cite{sac1, discrete-sac}.  The agent's observation comprises three stacked semantic segmentation frames, each with dimensions $84 \times 180$.  These frames may be presented with or without the incorporation of virtual guidance. The agent operates within an action space defined by a set $\mathcal{A}$. This set comprises two primary actions: $\textbf{NOOP}$ and $\textbf{TURN}(\alpha)$. Under the $\textbf{NOOP}$ action, the agent maintains its current directional orientation and travels along a straight trajectory. On the other hand, the $\textbf{TURN}(\alpha)$ action allows the agent to incrementally adjust its orientation based on the value $\alpha$. The sign of $\alpha$ is essential: negative values induce a leftward adjustment, while positive values prompt a rightward shift. This highlights the agent's ability to navigate and turn in a non-binary fashion. The angular velocity $\omega$, influenced by these continuous adjustments to $\alpha$, is formulated as follows:
\begin{equation}
    \omega \pluseq \alpha \times \kappa \times \Delta t,
\end{equation}
where $\Delta t$ represents the time interval, and $\kappa$ defines the steering sensitivity. In our settings, $\alpha$ has a standard value of $\SI{35}{\degree/\second^2}$, and $\kappa$ is set to two. Rather than abrupt binary switches in direction, our agent's trajectory evolves based on the cumulative impact of successive $\alpha$ adjustments. While the agent maintains a consistent velocity $v$ of $\SI{6}{\meter/\second}$, its direction continuously experiences refined and gradual modifications. It receives a reward of $10.0$ when reaching the destination, and a penalty of $-10.0$ for colliding with an obstacle, venturing outside the boundaries, or exceeding the time limit.

\input{tables/experiment1.tex}
\input{tables/experiment1-error-analysis.tex}

\subsubsection{Evaluation Metrics}
\label{subsec::evaluation_metric}
% \del{In our experiments, we employ multiple metrics to evaluate the agent's efficacy in utilizing forecasted information derived from different strategies and settings. These evaluation metrics are described as follows.}
\paragraph{Success rate}
In our evaluations, the success rate serves as a metric to assess the proficiency of the agent in reaching the designated destination.
% \del{Although maximizing the success rate may be achieved through exhaustive exploration, SPL introduces a balance by comparing the agent's success rate to its trajectory length.} 
\paragraph{Success rate weighted by path length (SPL)}
The SPL metric evaluates the agent's navigational performance by accounting for both the success in reaching the destination and the efficiency of the selected trajectory~\cite{SPL}. This comparison ensures a consideration of both navigational success and efficiency. SPL is mathematically represented as follows:
\begin{equation}
    \frac{1}{N} \sum_{i=1}^{N} \frac{l_i}{\max(l_i, p_i)},
\end{equation}
where $l_i$ represents the shortest path distance from the agent’s starting position to the goal for episode $i$, and $p_i$ denotes the length of the path actually taken by the agent in that episode.

\paragraph{Collision rate and out-of-bound (OOB) rate}
The collision and the out-of-bound rate are utilized as our metrics to analyze the causes of failure cases. Each failure case can be attributed to one of two possible causes: (a) \textit{out-of-bound}, which indicates the agent has ventured into prohibited regions such as the sidewalk, and (b) \textit{collision}, which represents instances where the agent collides with obstacles. % \del{in the environment, including static and dynamic objects.}

\paragraph{Final displacement error (FDE) and average displacement error (ADE)} ADE is defined as the mean Euclidean distance between the predicted and the ground-truth future positions. FDE, on the other hand, is defined in a similar manner but is calculated only for the final timestep.

\subsubsection{Hyperparameters}
In this study, the input resolution for object detection and segmentation models are set to $360 \times 640$ and $720 \times 1280$, respectively. We forecast bounding boxes for the future five time intervals, where each interval comprises four skipped frames.  For each experiment, we employ three independent and identically distributed (i.i.d.) random seeds. 

\subsubsection{Baseline Representations}
In this study, two baseline representations are considered for comparison: \textit{Seg} and \textit{Seg (box)}. The \textit{Seg (box)} representation extends the pedestrian segment to its bounding box, to enable a fair comparison with \textit{seg (box)} + \textit{BOX} to evaluate the efficacy of visual forecasting. This representation is particularly relevant given the enlarged coverage that bounding boxes provide for individual pedestrians. While \textit{Seg} captures the detailed contours of pedestrians, neither \textit{BOX} nor \textit{Seg (box)} offer this level of granularity.

\subsubsection{Hardware Configuration}
Our real-world experiments were carried out on a laptop equipped with an NVIDIA GeForce RTX 4080 Mobile GPU and an Intel Core i9 CPU. The robot platform is a ClearPath Husky Unmanned Ground Vehicle (UGV). The entire pipeline, encompassing object detection, tracking, and segmentation, achieved an inference time of 0.03 seconds (i.e., $\sim 33$ frames per second (FPS)).

\subsection{Validation of the Effectiveness of Visual Forecasting}
\label{subsec:visual_forecasting_can_help}
In this section, we focus on validating the effectiveness of visual forecasting, with a particular emphasis on its role in enhancing the performance of DRL agents through CVM. Our evaluations are conducted using the \textit{S-Turn} environment, as the primary objective is to assess the agent's object avoidance capabilities. Evaluations of the other forecasting strategies will be discussed in the next section. Table~\ref{tab:visual-forecasting-effectness} presents the evaluation results. The visualized demonstrations of the forecasted trajectories, along with the agents' interactions, are illustrated in Fig.~\ref{fig:exp-virtual}.  The results suggest that the agents equipped with visual forecasting consistently outperform those without this feature in terms of both the \textit{SPL} and \textit{success rate} metrics under two different visual forecasting configurations.  These findings indicate that visual forecasting is able to provide informative foresight into the potential trajectories of dynamic objects, and enables the agents to plan ahead as well as avoiding potential collisions.

To further interpret the implications of visual forecasting, we undertake an analytical examination of the failure cases to identify their root causes. As evidenced in Table~\ref{tab:visual-forecasting-error-analysis}, the agents trained with visual forecasting exhibit a lower \textit{collision rate} in comparison to their counterparts trained without visual forecasting. This suggests the efficacy of visual forecasting in aiding the agents to interpret the motions of dynamic objects.

\input{tables/experiment2.tex}

\begin{figure}[t]
	\centering
    \includegraphics[width=\linewidth]{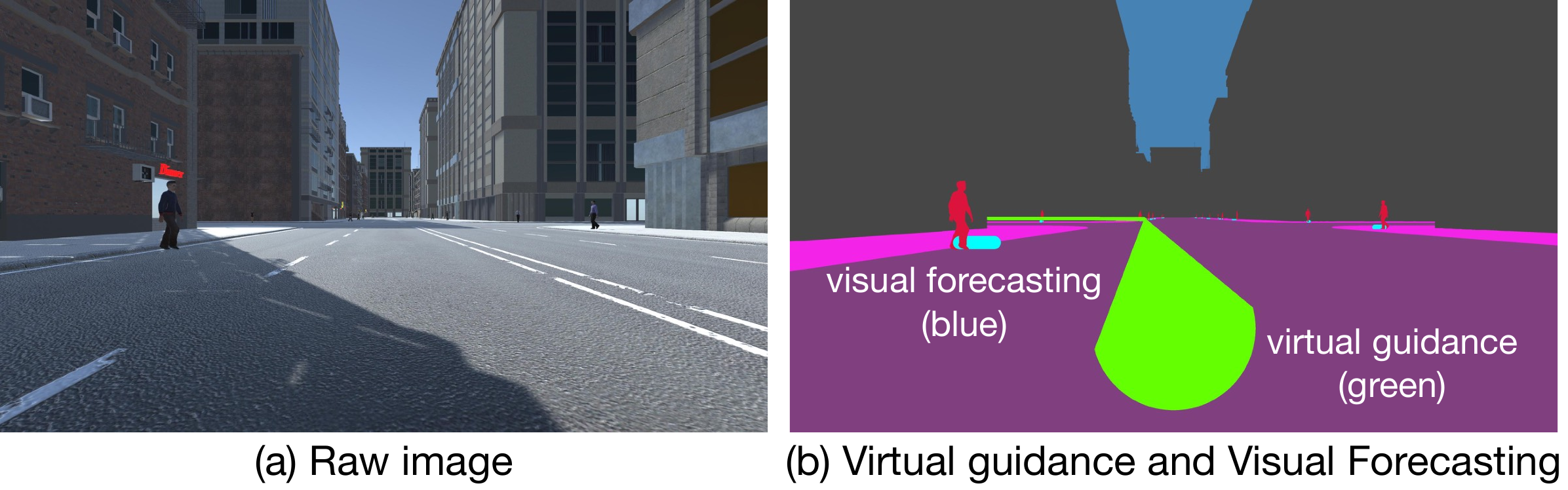}
    \caption{Visualization of virtual guidance~\cite{yang2023vision} and visual forecasting.}
    \label{fig:virtual-guidance}
\end{figure}

\input{tables/experiment3.tex}

\subsection{Quality of Forecasting: Is Constant Velocity Sufficient?}
\label{subsec:quality_of_forecasting}
To demonstrate the significance of visual forecasting, the previous section highlighted that the agents, even when guided by the most straightforward CVM forecasting approach, outperform those without visual forecasting.  While the accuracy of visual forecasting estimations might be pertinent to certain downstream tasks and high forecasting accuracy is often desired, its presence does not always signify improved outcomes in navigation and obstacle avoidance. As a result, this section investigates the relationship between forecasting accuracy and the agent's capabilities in navigation and object avoidance. For this examination, we utilize both CVM and KF as forecasting mechanisms, and compare their outcomes with the Ground Truth (GT) forecasting as the benchmark for ideal future location predictions characterized by zero FDE and ADE errors. Our evaluation is performed on the \textit{S-Turn} environment and encompasses two distinct perspectives: (a) the precision of forecasting, measured by FDE and ADE over five steps across 100 pedestrian samples, and (b) the proficiency of the agent, reflected through the metrics of \textit{SPL} and \textit{success rate}. The results are presented in Table~\ref{table:forecasting-quality}. It can be observed that the forecasting accuracy of the CVM closely approximates that of the KF. Both models are able to deliver satisfactory results, with only slight margins compared to the perfect predictions of the GT forecasting. This finding aligns with the prior studies in~\cite{Schller2019WhatTC, wu2023truly, uhlemann2023evaluating}. The results therefore suggest the efficacy of even a CVM in producing reliable forecasts.

Table~\ref{table:forecasting-quality} provides further validation regarding the influence of forecasting quality on an agent's ability in object avoidance. As indicated by the SPL metric in Table~\ref{table:forecasting-quality}, even basic models such as CVM or KF offer beneficial insights to DRL agents in object avoidance tasks. It is worth noting that the agents trained with forecasts from CVM and KF display performance levels nearing those trained with the GT's impeccable visual forecasting.  The explanation to this observation could be that the primary advantage of visual forecasting lies in its capability to offer an intuitive depiction of general trajectories. Even without intricate details, these visual indications prove to be highly effective in directing agents. The combined simplicity and efficacy of CVM suggest its potential for adaptation and practical use in real-world situations, given its low computational requirements.

\subsection{Compatibility of Visual Forecasting with Navigation}
\label{subsec:visual_forecasting_with_virtual_navigation}
In this section, we investigate the interplay between visual forecasting and the virtual guidance-based navigation technique introduced in~\cite{yang2023vision}. Building on the premise established in the previous sections that visual forecasting can assist agents, the \textit{Urban Grid Street} environment is selected for this examination.  This environment serves as a more generalized setting and allows us to adopt a diverse combination of routes to assess the robustness and adaptability of the forecasting models. The integration of these two methods is depicted in Fig.~\ref{fig:virtual-guidance}. The evaluation results are presented in Table~\ref{tables:virtual-guidance-evaluation}. It can be observed that the agents trained with visual forecasting consistently outperform those without it, as evidenced by the \textit{SPL} and \textit{success rate} metrics. This not only highlights the complementary nature of visual forecasting and virtual navigation schemes, but also reveals the DRL agents' ability to comprehend information from these two different methods.

\begin{figure}[t]
  \centering
  \includegraphics[width=\linewidth]{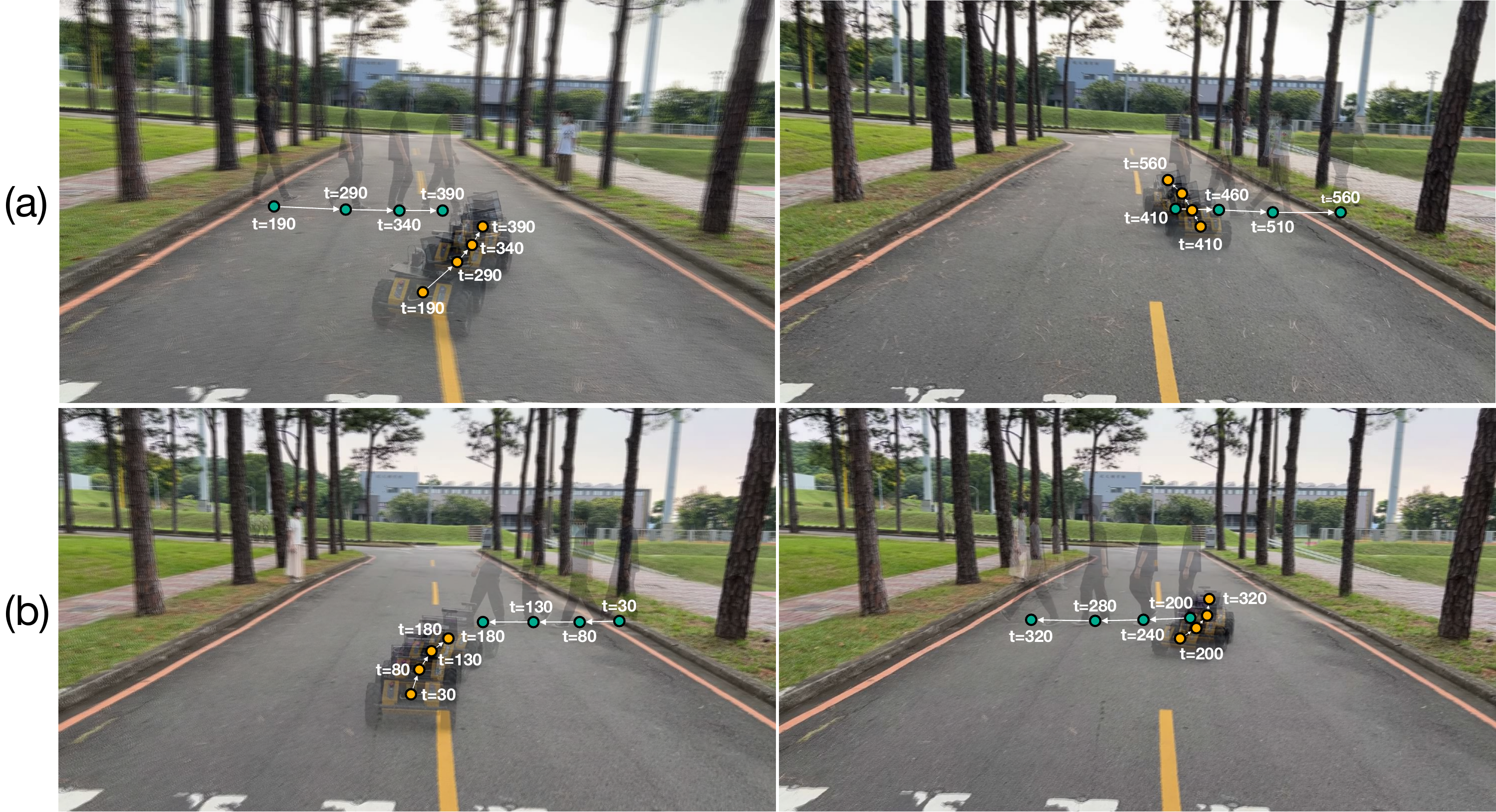}
  \caption{Demonstration of the agent's performance in real-world scenarios.}
  \label{fig:exp-real}
\end{figure}

\subsection{Real World Transferring}
In this section, we evaluate the trained agent's performance in real-world scenarios to understand the potential of visual forecasting as a mid-level representation and its efficacy in aiding agents with object avoidance tasks. A consideration prior to this transition is the agent's reliance on 2D information extracted from RGB images, as discussed in Section~\ref{subsubsec::forecast_real_world}, especially in the absence of precise 3D information about objects. To circumvent these limitations, we introduce a modified approach that relies solely on predicted 2D bounding boxes for visual forecasting. The results, presented in Table~\ref{tab:2d-forecasting}, demonstrate the adaptability of our method in real-world settings. It is evident that using even just 2D bounding boxes can improve the agent's performance compared to not using visual forecasting at all, although the effectiveness is somewhat less than when utilizing 3D bounding boxes. Given that 3D prediction models could demand more computational resources and are not the primary focus of this paper, these findings serve as a preliminary indication that satisfactory performance can be achieved even with 2D bounding boxes.

The results of our real-world experiments are presented in Table~\ref{tab:comparison}. The agents are evaluated both with and without the incorporation of visual forecasting. Each experiment is conducted over twenty independent runs, and the task involves guiding the agent to a predetermined destination while avoiding collisions with three moving pedestrians. We use the \textit{completion rate} metric to evaluate the agent's ability to dodge pedestrians successfully. It can be observed that the agents equipped with visual forecasting (i.e., \textit{Seg} + \textit{AP} and \textit{Seg~(box)} + \textit{BOX}) outperform those without forecasting. Fig.~\ref{fig:exp-real} further illustrates this by showcasing two examples of the agent behavior using the \textit{Seg} + \textit{AP} forecasting scheme. In both instances, the agents appear to take into account the future positions of the pedestrians when making decisions.

\section{Conclusion}
\label{sec:conclusion}
In this paper, we addressed the challenges of navigation and obstacle avoidance for autonomous agents in dynamic environments. We introduced visual forecasting as an innovative strategy that used intuitive visual cues to predict the future paths of dynamic objects. Our method employed mid-level representations to translate anticipated future trajectories into easily digestible formats. We proposed two schemes: sequences of bounding boxes and augmented paths. To validate our approach, we conducted experiments in both virtual and real-world settings. Our experimental results confirmed the effectiveness and applicability of visual forecasting in aiding navigation and obstacle avoidance tasks.

%===============================================================================

\bibliographystyle{IEEEtran}
\bibliography{reference}

\end{document}

%% file: tables/experiment1.tex
\begin{table*}[t]
\caption{A comparison for showing the beneficial impact of visual forecasting.}% \del{\textit{AP.} and \textit{BOX.} denotes the visual forecasting represented by augmented path and boudning box, respectively.}} 
\label{tab:visual-forecasting-effectness}
\centering
\newcommand{\xmark}{\ding{55}}
\resizebox{\linewidth}{!}{%
\renewcommand{\arraystretch}{1.3}
\begin{tabular}{ c | c | rlrlrl | cccccc }
\toprule     
\multirow{2}{*}{\textbf{Approaches}} & \multirow{2}{*}{\parbox[c]{1.5cm}{\centering\textbf{Visual} \textbf{Forecasting}}}  & \multicolumn{6}{c|}{\textbf{Success Rate Weighted by Path Length (SPL)}}  & \multicolumn{6}{c}{\textbf{Success Rate}} \\
 & & \multicolumn{2}{c}{$[\SI{0.3}{m/\second}$, $\SI{0.6}{m/\second}]$} & \multicolumn{2}{c}{$[\SI{0.6}{m/\second}$, $\SI{1.2}{m/\second}]$} & 
 \multicolumn{2}{c|}{$[\SI{1.2}{m/\second}$, $\SI{1.5}{m/\second}]$} &
 \multicolumn{2}{c}{$[\SI{0.3}{m/\second}$, $\SI{0.6}{m/\second}]$} & 
 \multicolumn{2}{c}{$[\SI{0.6}{m/\second}$, $\SI{1.2}{m/\second}]$} & 
 \multicolumn{2}{c}{$[\SI{1.2}{m/\second}$, $\SI{1.5}{m/\second}]$} \\ 
\midrule
 \leftcelll{\textit{Seg (box)}} & \xmark & $\SI{82.14}{\percent}$ & \multirow{2}{*}{\textcolor{teal}{($\uparrow$ $\SI{2.32}{\percent}$)}} & $\SI{68.47}{\percent}$ & \multirow{2}{*}{\textcolor{teal}{($\uparrow$ $\SI{11.77}{\percent}$)}} & $\SI{47.55}{\percent}$ & \multirow{2}{*}{\textcolor{teal}{($\uparrow$ $\SI{25.89}{\percent}$)}} & $\SI{88.25}{\percent}$ & \multirow{2}{*}{\textcolor{teal}{($\uparrow$ $\SI{4.27}{\percent}$)}} & $\SI{74.27}{\percent}$ &  \multirow{2}{*}{\textcolor{teal}{($\uparrow$ $\SI{14.16}{\percent}$)}} & $\SI{51.91}{\percent}$ & \multirow{2}{*}{\textcolor{teal}{($\uparrow$ $\SI{29.59}{\percent}$)}} \\
 \leftcelll{\textit{Seg (box)} + \textit{BOX}}  & \checkmark & \underline{$\SI{83.86}{\percent}$} & & \underline{$\SI{81.67}{\percent}$} & & \underline{$\SI{75.83}{\percent}$} &  & \underline{$\SI{92.03}{\percent}$} & & \leftcell{\underline{$\SI{89.67}{\percent}$}} & & \leftcell{\underline{$\SI{83.63}{\percent}$}} & \\ 
\midrule
 \leftcelll{\textit{Seg}} & \xmark & $\SI{79.94}{\percent}$ & \multirow{2}{*}{\textcolor{teal}{($\uparrow \SI{6.20}{\percent}$)}} & $\SI{62.28}{\percent}$ & \multirow{2}{*}{\textcolor{teal}{($\uparrow$ $\SI{19.61}{\percent}$)}} & $\SI{39.88}{\percent}$ & \multirow{2}{*}{\textcolor{teal}{($\uparrow$ $\SI{35.12}{\percent}$)}} & $\SI{85.41}{\percent}$ & \multirow{2}{*}{\textcolor{teal}{($\uparrow$ $\SI{6.33}{\percent}$)}} & $\SI{67.29}{\percent}$ & \multirow{2}{*}{\textcolor{teal}{($\uparrow$ $\SI{20.45}{\percent}$)}} & \leftcell{$\SI{43.29}{\percent}$} & \multirow{2}{*}{\textcolor{teal}{($\uparrow$ $\SI{37.63}{\percent}$)}} \\
 \leftcelll{\textit{Seg} + \textit{AP}} & \checkmark & \underline{$\SI{86.14}{\percent}$} & & \underline{$\SI{81.89}{\percent}$} & & \underline{$\SI{75.00}{\percent}$} & & \underline{$\SI{91.74}{\percent}$} & & \underline{$\SI{87.74}{\percent}$} & & \underline{$\SI{80.92}{\percent}$} \\ 
\bottomrule
\end{tabular}}
\end{table*}

%% file: tables/experiment1-error-analysis.tex
\begin{table}[t]
\caption{An Analysis of failure cases.}
\label{tab:visual-forecasting-error-analysis}
\centering
\newcommand{\xmark}{\ding{55}}
\resizebox{\linewidth}{!}{
\renewcommand{\arraystretch}{1.3}
\begin{tabular}{c | c | cc}
\toprule
\multirow{2}{*}{\textbf{Approaches}} & \multirow{2}{*}{\textbf{Visual Forecasting}} & \multicolumn{2}{c}{\textbf{Causes of the Failure Cases}}\\ 
& & \textit{Collision Rate ($\downarrow$)} & \textit{Out-Of-Bound Rate ($\downarrow$)} \\ 
\midrule
\leftcelll{\textit{Seg (box)}} & \xmark & $\SI{24.19}{\percent}$ & $\SI{4.33}{\percent}$  \\
\leftcelll{\textit{Seg (box)} + \textit{BOX}}  & \checkmark & \underline{$\SI{7.34}{\percent}$} & \underline{$\SI{4.22}{\percent}$} \\ 
\midrule
\leftcelll{\textit{Seg}} & \xmark & $\SI{29.40}{\percent}$ & $\SI{5.27}{\percent}$  \\
\leftcelll{\textit{Seg} + \textit{AP}} & \checkmark & \underline{$\SI{9.73}{\percent}$} & \underline{$\SI{3.48}{\percent}$} \\
\bottomrule
\end{tabular}}
\end{table}

%% file: tables/experiment2.tex
\begin{table}[t]
\caption{The evaluation of the forecasting quality.} 
\label{table:forecasting-quality}
\centering
\newcommand{\xmark}{\ding{55}}
\resizebox{\linewidth}{!}{%
\renewcommand{\arraystretch}{1.2}
\begin{tabular}{ c | c | c | c | c | c }
\toprule     
 \textbf{Visual Forecasting} & \textbf{Algorithm} & \textbf{FDE ($\downarrow$)} & \textbf{ADE ($\downarrow$)} & \textbf{SPL ($\uparrow$)}  & \textbf{Success Rate ($\uparrow$)} \\
\midrule
 \multirow{4}{*}{\textit{BOX}}
                   & \xmark      & \NA & \NA & $\SI{66.05}{\percent}$ & $\SI{71.48}{\percent}$ \\
                   & \textit{CVM} & $\SI{0.471}{}$ & $\SI{0.283}{}$ & $\SI{80.45}{\percent}$ & $\SI{88.44}{\percent}$ \\
                   & \textit{KF} & $\SI{0.456}{}$ & $\SI{0.340}{}$ & $\SI{82.71}{\percent}$ & $\SI{89.22}{\percent}$ \\ 
                   & \textit{GT} & \NA & \NA & $\SI{83.85}{\percent}$ & $\SI{90.83}{\percent}$ \\
                   
\midrule
 \multirow{4}{*}{\textit{AP}} 
                   & \xmark      & \NA & \NA & $\SI{60.70}{\percent}$ & $\SI{65.33}{\percent}$ \\
                   & \textit{CVM} & $\SI{0.471}{}$ & $\SI{0.283}{}$ & $\SI{81.01}{\percent}$ & $\SI{86.80}{\percent}$ \\
                   & \textit{KF} & $\SI{0.456}{}$ & $\SI{0.340}{}$ & $\SI{81.63}{\percent}$ & $\SI{87.27}{\percent}$ \\ 
                   & \textit{GT} & \NA & \NA & $\SI{83.53}{\percent}$ & $\SI{87.90}{\percent}$ \\
\bottomrule
\end{tabular}}
\end{table}

%% file: tables/experiment3.tex
\begin{table*}[t]
\centering
\newcommand{\xmark}{\ding{55}}
\begin{minipage}[t]{.28\linewidth}
\renewcommand{\arraystretch}{1.7}
\caption{The results of visual forecasting with virtual navigation.} 
\label{tables:virtual-guidance-evaluation}
\resizebox{\linewidth}{!}{%
\begin{tabular}[t]{c | c | c | c}
\toprule     
\textbf{Scenario}                & \textbf{Approaches}            & \textbf{SPL} & \textbf{Success Rate} \\
\midrule
\multirow{2}{*}{\textbf{seen}}  
                                 & \leftcelll{\textit{Seg}}       &  $\SI{65.60}{\percent}$   & $\SI{65.78}{\percent}$ \\
                                 & \leftcelll{\textit{Seg + AP}}  &  \underline{$\SI{77.92}{\percent}$}   & \underline{$\SI{78.19}{\percent}$} \\ \midrule
\multirow{2}{*}{\textbf{unseen}} 
                                 & \leftcelll{\textit{Seg}}       &  $\SI{47.06}{\percent}$   & $\SI{47.50}{\percent}$ \\
                                 & \leftcelll{\textit{Seg + AP}}  &  \underline{$\SI{63.30}{\percent}$}   & \underline{$\SI{63.83}{\percent}$} \\ \bottomrule
\end{tabular}}
\end{minipage} \hfill
\begin{minipage}[t]{.39\linewidth}
\renewcommand{\arraystretch}{1.0}
\caption{Comparison of utilizing 2D and 3D bounding boxes for performing visual forecasting.}
\label{tab:2d-forecasting}
\resizebox{\linewidth}{!}{%
\begin{tabular}[t]{c | c | cc}
\toprule
\textbf{Approaches}                                 & \textbf{Require 3D?}                                                                  & \textbf{SPL} & \textbf{Success Rate}\\ \midrule
\leftcelll{\textit{Seg (box)}}                      & \NA                                                                                   & $\SI{66.05}{\percent}$    & $\SI{71.48}{\percent}$             \\
\leftcelll{\textit{Seg (box)} + \textit{BOX (2D)}}  & \xmark                                                                                & $\SI{71.65}{\percent}$    & $\SI{77.33}{\percent}$             \\ 
\leftcelll{\textit{Seg (box)} + \textit{BOX (3D)}}  & \checkmark                                                                            & $\SI{79.38}{\percent}$    & $\SI{87.48}{\percent}$             \\ \midrule
\leftcelll{\textit{Seg}}                            & \NA                                                                                   & $\SI{60.70}{\percent}$    & $\SI{65.33}{\percent}$             \\
\leftcelll{\textit{Seg} + \textit{AP (2D)}}         & \xmark                                                                                & $\SI{71.89}{\percent}$    & $\SI{78.00}{\percent}$             \\ 
\leftcelll{\textit{Seg} + \textit{AP (3D)}}         & \checkmark                                                                            & $\SI{81.01}{\percent}$    & $\SI{86.80}{\percent}$             \\ \bottomrule
\end{tabular}}
\end{minipage} \hfill
\begin{minipage}[t]{.3\linewidth}
\renewcommand{\arraystretch}{1.5}
\label{tab:comparison}
\caption{The evaluation results in the real-world environment.} 
\resizebox{\linewidth}{!}{%
\begin{tabular}[t]{c | cc }
\toprule
    \textbf{Approaches}                              &  \textbf{Completion Rate} & \textbf{Success Rate} \\ \midrule 
    \leftcelll{\textit{Seg (box)}}                  & $\SI{20.00}{\percent}$                 & $\SI{5.00}{\percent}$              \\
    \leftcelll{\textit{Seg (box)} + \textit{BOX}}   & \underline{$\SI{48.33}{\percent}$}     & \underline{$\SI{25.00}{\percent}$} \\ \midrule
    \leftcelll{\textit{Seg}}                        & $\SI{31.67}{\percent}$                 & $\SI{5.00}{\percent}$              \\
    \leftcelll{\textit{Seg} + \textit{AP}}          & \underline{$\SI{73.33}{\percent}$}     & \underline{$\SI{60.00}{\percent}$} \\ 
\bottomrule
\end{tabular}}
\end{minipage}
\end{table*}

%% file: root.bbl
\begin{thebibliography}{10}
\providecommand{\url}[1]{#1}
\csname url@rmstyle\endcsname
\providecommand{\newblock}{\relax}
\providecommand{\bibinfo}[2]{#2}
\providecommand\BIBentrySTDinterwordspacing{\spaceskip=0pt\relax}
\providecommand\BIBentryALTinterwordstretchfactor{4}
\providecommand\BIBentryALTinterwordspacing{\spaceskip=\fontdimen2\font plus
\BIBentryALTinterwordstretchfactor\fontdimen3\font minus
  \fontdimen4\font\relax}
\providecommand\BIBforeignlanguage[2]{{%
\expandafter\ifx\csname l@#1\endcsname\relax
\typeout{** WARNING: IEEEtran.bst: No hyphenation pattern has been}%
\typeout{** loaded for the language `#1'. Using the pattern for}%
\typeout{** the default language instead.}%
\else
\language=\csname l@#1\endcsname
\fi
#2}}

\bibitem{LIU2019272}
Z.~Liu, Q.~Liu, W.~Xu, Z.~Liu, Z.~Zhou, and J.~Chen, ``Deep learning-based
  human motion prediction considering context awareness for human-robot
  collaboration in manufacturing,'' \emph{Procedia CIRP}, vol.~83, pp.
  272--278, 2019.

\bibitem{poddar2023crowd}
S.~Poddar, C.~Mavrogiannis, and S.~S. Srinivasa, ``From crowd motion prediction
  to robot navigation in crowds,'' \emph{arXiv preprint arXiv:2303.01424},
  2023.

\bibitem{wang22e}
A.~Wang, C.~Mavrogiannis, and A.~Steinfeld, ``Group-based motion prediction for
  navigation in crowded environments,'' in \emph{Proceedings of Conference on
  Robot Learning (CoRL)}, 2022, pp. 871--882.

\bibitem{Park_2016_IROS}
C.~Park, J.~Ondřej, M.~Gilbert, K.~Freeman, and C.~O'Sullivan, ``Hi robot:
  Human intention-aware robot planning for safe and efficient navigation in
  crowds,'' in \emph{Proceedings of IEEE/RSJ International Conference on
  Intelligent Robots and Systems (IROS)}, 2016, pp. 3320--3326.

\bibitem{Chen2018RobotNB}
Z.~Chen, C.~Song, Y.~Yang, B.~Zhao, Y.~Hu, S.~B. Liu, and J.~Zhang, ``Robot
  navigation based on human trajectory prediction and multiple travel modes,''
  \emph{Applied Sciences}, 2018.

\bibitem{Fraichard2020FromCS}
T.~Fraichard and V.~Levesy, ``From crowd simulation to robot navigation in
  crowds,'' \emph{IEEE Robotics and Automation Letters}, 2020.

\bibitem{trajectory-pred-for-avoidance}
A.~Feher, S.~Aradi, and T.~Becsi, ``Online trajectory planning with
  reinforcement learning for pedestrian avoidance,'' \emph{Electronics}, 2022.

\bibitem{Mavrogiannis2021WindingTC}
C.~Mavrogiannis, K.~Balasubramanian, S.~Poddar, A.~Gandra, and S.~S. Srinivasa,
  ``Winding through: Crowd navigation via topological invariance,'' \emph{IEEE
  Robotics and Automation Letters}, 2021.

\bibitem{Hong2018IJCAI}
Z.-W. Hong, Y.-M. Chen, H.-K. Yang, S.-Y. Su, T.-Y. Shann, Y.-H. Chang,
  B.~H.-L. Ho, C.-C. Tu, T.-C. Hsiao, H.-W. Hsiao, S.-P. Lai, Y.-C. Chang, and
  C.-Y. Lee, ``Virtual-to-real: Learning to control in visual semantic
  segmentation,'' in \emph{Proceedings of the International Joint Conference on
  Artificial Intelligence (IJCAI)}, 2018, pp. 4912--4920.

\bibitem{zhao2020sim2real}
W.~Zhao, J.~P. Queralta, and T.~Westerlund, ``Sim-to-real transfer in deep
  reinforcement learning for robotics: a survey,'' in \emph{SSCI}, 2020.

\bibitem{Lin2020LearningTS}
Y.-C. Lin, A.~Zeng, S.~Song, P.~Isola, and T.-Y. Lin, ``Learning to see before
  learning to act: Visual pre-training for manipulation,'' in \emph{Proceedings
  of IEEE International Conference on Robotics and Automation (ICRA)}, 2020,
  pp. 7286--7293.

\bibitem{chen_2020_CORL}
B.~Chen, A.~Sax, F.~Lewis, I.~Armeni, S.~Savarese, A.~Zamir, J.~Malik, and
  L.~Pinto, ``Robust policies via mid-level visual representations: An
  experimental study in manipulation and navigation,'' in \emph{Proceedings of
  the Conference on Robot Learning (CoRL)}, 2021, pp. 2328--2346.

\bibitem{unity-eng}
{\relax Unity Technologies}, ``Unity engine,'' https://unity.com.

\bibitem{yang2023vision}
H.-K. Yang \emph{et~al.}, ``Vision based virtual guidance for navigation,''
  \emph{arXiv preprint arXiv:2303.02731}, 2023.

\bibitem{Yang2022IROS}
H.-K. Yang, T.-C. Hsiao, T.-H. Liao, H.-S. Liu, L.-Y. Tsao, T.-W. Wang, S.-Y.
  Yang, Y.-W. Chen, H.-R. Liao, and C.-Y. Lee, ``Investigation of factorized
  optical flows as mid-level representations,'' in \emph{Proceedings of
  IEEE/RSJ International Conference on Intelligent Robots and Systems (IROS)},
  2022, pp. 746--753.

\bibitem{Gupta_2018_CVPR}
A.~Gupta, J.~Johnson, L.~Fei-Fei, S.~Savarese, and A.~Alahi, ``Social gan:
  Socially acceptable trajectories with generative adversarial networks,'' in
  \emph{Proceedings of the IEEE Conference on Computer Vision and Pattern
  Recognition (CVPR)}, June 2018.

\bibitem{Salzmann_2020_TrajectronDT}
T.~Salzmann, B.~Ivanovic, P.~Chakravarty, and M.~Pavone, ``Trajectron++:
  Dynamically-feasible trajectory forecasting with heterogeneous data,'' in
  \emph{Proceedings of the European Conference on Computer Vision (ECCV)},
  2020.

\bibitem{Mangalam_2020_WACV}
K.~Mangalam, E.~Adeli, K.-H. Lee, A.~Gaidon, and J.~C. Niebles, ``Disentangling
  human dynamics for pedestrian locomotion forecasting with noisy
  supervision,'' in \emph{Proceedings of the IEEE/CVF Winter Conference on
  Applications of Computer Vision (WACV)}, March 2020.

\bibitem{Yuan_2021_ICCV}
Y.~Yuan, X.~Weng, Y.~Ou, and K.~M. Kitani, ``Agentformer: Agent-aware
  transformers for socio-temporal multi-agent forecasting,'' in
  \emph{Proceedings of the IEEE/CVF International Conference on Computer Vision
  (ICCV)}, October 2021, pp. 9813--9823.

\bibitem{Pang_2021_CVPR}
B.~Pang, T.~Zhao, X.~Xie, and Y.~N. Wu, ``Trajectory prediction with latent
  belief energy-based model,'' in \emph{Proceedings of the IEEE/CVF Conference
  on Computer Vision and Pattern Recognition (CVPR)}, June 2021, pp.
  11\,814--11\,824.

\bibitem{Shafiee_2021_CVPR}
N.~Shafiee, T.~Padir, and E.~Elhamifar, ``Introvert: Human trajectory
  prediction via conditional 3d attention,'' in \emph{Proceedings of the
  IEEE/CVF Conference on Computer Vision and Pattern Recognition (CVPR)}, June
  2021, pp. 16\,815--16\,825.

\bibitem{unsup-interaction-video-prediction}
C.~Finn, I.~Goodfellow, and S.~Levine, ``Unsupervised learning for physical
  interaction through video prediction,'' in \emph{Advances in Neural
  Information Processing Systems}, 2016.

\bibitem{PiSAC}
K.-H. Lee, I.~Fischer, A.~Z. Liu, Y.~Guo, H.~Lee, J.~Canny, and S.~Guadarrama,
  ``Predictive information accelerates learning in rl,'' in \emph{Advances in
  Neural Information Processing Systems}, 2020.

\bibitem{visualforesight}
F.~Ebert, C.~Finn, S.~Dasari, A.~Xie, A.~Lee, and S.~Levine, ``Visual
  foresight: Model-based deep reinforcement learning for vision-based robotic
  control,'' \emph{arXiv preprint arXiv:1812.00568}, 2018.

\bibitem{Zeng_2020_CVPR}
K.-H. Zeng, R.~Mottaghi, L.~Weihs, and A.~Farhadi, ``Visual reaction: Learning
  to play catch with your drone,'' in \emph{IEEE/CVF Conference on Computer
  Vision and Pattern Recognition (CVPR)}, June 2020.

\bibitem{kalmanfilter}
``A new approach to linear filtering and prediction problems,'' vol.~82, 1960,
  pp. 35--45.

\bibitem{aharon2022bot}
N.~Aharon, R.~Orfaig, and B.-Z. Bobrovsky, ``Bot-sort: Robust associations
  multi-pedestrian tracking,'' \emph{arXiv preprint arXiv:2206.14651}, 2022.

\bibitem{Tranheden_2021_WACV}
W.~Tranheden, V.~Olsson, J.~Pinto, and L.~Svensson, ``Dacs: Domain adaptation
  via cross-domain mixed sampling,'' in \emph{Proceedings of the IEEE/CVF
  Winter Conference on Applications of Computer Vision (WACV)}, January 2021,
  pp. 1379--1389.

\bibitem{Wojke2017simple}
N.~Wojke, A.~Bewley, and D.~Paulus, ``Simple online and realtime tracking with
  a deep association metric,'' in \emph{Proceedings of IEEE International
  Conference on Image Processing (ICIP)}, 2017, pp. 3645--3649.

\bibitem{zhang2022bytetrack}
Y.~Zhang, P.~Sun, Y.~Jiang, D.~Yu, F.~Weng, Z.~Yuan, P.~Luo, W.~Liu, and
  X.~Wang, ``Bytetrack: Multi-object tracking by associating every detection
  box,'' 2022.

\bibitem{sac1}
T.~Haarnoja \emph{et~al.}, ``Soft actor-critic: Off-policy maximum entropy deep
  reinforcement learning with a stochastic actor,'' in \emph{ICML}, 2018.

\bibitem{discrete-sac}
P.~Christodoulou, ``Soft actor-critic for discrete action settings,''
  \emph{arXiv preprint arXiv:1910.07207}, 2019.

\bibitem{SPL}
P.~Anderson, A.~X. Chang, D.~S. Chaplot, A.~Dosovitskiy, S.~Gupta, V.~Koltun,
  J.~Kosecka, J.~Malik, R.~Mottaghi, M.~Savva, and A.~R. Zamir, ``On evaluation
  of embodied navigation agents,'' \emph{arXiv preprint arXiv:1807.06757}.

\bibitem{Schller2019WhatTC}
C.~Sch{\"o}ller, V.~Aravantinos, F.~S. Lay, and A.~Knoll, ``What the constant
  velocity model can teach us about pedestrian motion prediction,'' \emph{IEEE
  Robotics and Automation Letters}, vol.~5, pp. 1696--1703, 2019.

\bibitem{wu2023truly}
H.~Wu, T.~Phong, C.~Yu, P.~Cai, S.~Zheng, and D.~Hsu, ``What truly matters in
  trajectory prediction for autonomous driving?'' \emph{arXiv preprint
  arXiv:2306.15136}, 2023.

\bibitem{uhlemann2023evaluating}
N.~Uhlemann, F.~Fent, and M.~Lienkamp, ``Evaluating pedestrian trajectory
  prediction methods for the application in autonomous driving,'' \emph{arXiv
  preprint arXiv:2306.15136}, 2023.

\end{thebibliography}
